\UseRawInputEncoding
\documentclass[10pt,twocolumn,letterpaper]{article} 

\usepackage{caption}
\usepackage{subcaption}
\usepackage{avss}
\usepackage{times}
\usepackage{epsfig}
\usepackage{graphicx}
\usepackage{amsmath}
\usepackage{amssymb}
\usepackage[scientific-notation=true]{siunitx}
\usepackage{soul}

\usepackage[pagebackref=true,breaklinks=true,letterpaper=true,colorlinks,bookmarks=false]{hyperref}

\avssfinalcopy 

\def\avssPaperID{97} 

\ifavssfinal\fi

\makeatletter
\def\ps@IEEEtitlepagestyle{%
  \def\@oddfoot{\mycopyrightnotice}%
  \def\@evenfoot{}%
}
\def\mycopyrightnotice{%
  {\footnotesize 978-1-6654-3396-9/21/\$31.00 \copyright 2021 IEEE\hfill}
  \gdef\mycopyrightnotice{}
}
\makeatother

\begin{document}

\title{Sequence Models for Drone vs Bird Classification}

\author{Fatih C. Akyon$^{1,2}$, Erdem Akagunduz$^{2}$, Sinan O. Altinuc$^{1,2}$, Alptekin Temizel$^{2}$ \\ \\
$^{1}$OBSS AI, Ankara, Turkey \\
$^{2}$Graduate School of Informatics, METU, Ankara, Turkey \\
{\tt\small \{fatih.akyon,akaerdem,atemizel\}@metu.edu.tr, sinan.altinuc@obss.com.tr}
}

\maketitle
\thispagestyle{empty}

\begin{abstract}
Drone detection has become an essential task in object detection as drone costs have decreased and drone technology has improved. It is, however, difficult to detect distant drones when there is weak contrast, long range, and low visibility. In this work, we propose several sequence classification architectures to reduce the detected false-positive ratio of drone tracks. Moreover, we propose a new drone vs. bird sequence classification dataset to train and evaluate the proposed architectures. 3D CNN, LSTM, and Transformer based sequence classification architectures have been trained on the proposed dataset to show the effectiveness of the proposed idea. As experiments show, using sequence information, bird classification and overall F1 scores can be increased by up to 73\% and 35\%, respectively. Among all sequence classification models, R(2+1)D-based fully convolutional model yields the best transfer learning and fine-tuning results.
\end{abstract}

\section{Introduction}
Initially used for military purposes, drones are now used in a variety of other fields as well, such as traffic and weather monitoring \cite{elloumi2018monitoring}, and smart agriculture monitoring \cite{tokekar2016sensor}, just to name a few \cite{shakhatreh2019unmanned}. In addition, since the COVID-19 pandemic, drones have been increasingly used to deliver groceries and medical supplies autonomously and enforce social segregation. Nowadays, small quadcopters can be easily purchased on the Internet at low prices, which brings unprecedented opportunities but poses several threats regarding safety, privacy, and security \cite{humphreys2015statement}.

As a result of this recent and rapid interest on the drone utilization, the first Drone vs. Bird Detection Challenge was launched in 2017, in conjunction with the first edition of the International Workshop on Small-Drone Surveillance, Detection and Counteraction Techniques (WOSDETC)\footnote{This workshop was a part of the 14th edition of the IEEE International Conference on Advanced Video and Signal based Surveillance (AVSS).} \cite{coluccia2019drone}. This challenge aimed to address the technical issues of discriminating between drones and birds \cite{coluccia2019drone}. In practice, drones can be easily confused with birds, particularly at long distances, which makes the surveillance task even more challenging. The use of video analytics can solve the issue, but effective algorithms are needed, which can operate under unfavorable conditions, such as weak contrast, long-range, low visibility, etc.

\begin{figure}[!t]
\begin{center}
   \includegraphics[width=1\linewidth]{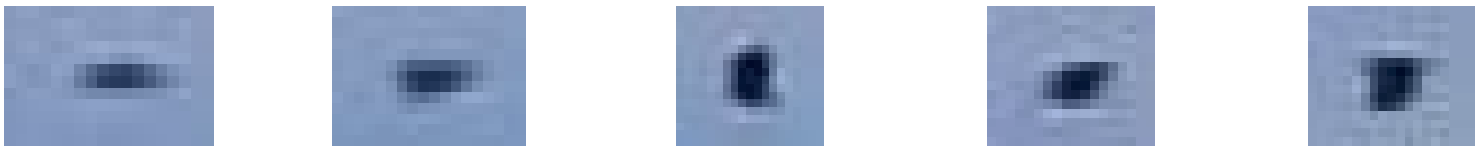}
\end{center}

\vspace{-0.3cm}

\begin{center}
   \includegraphics[width=1\linewidth]{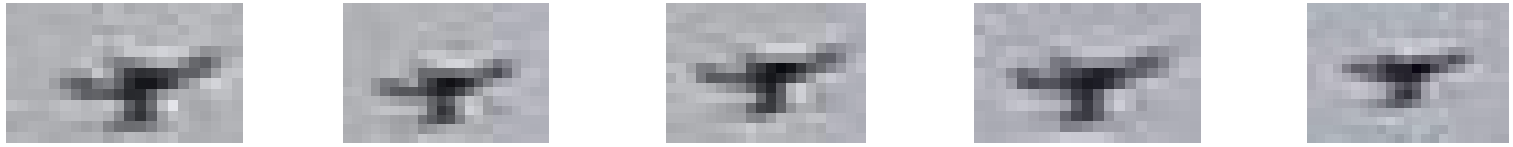}
\end{center}
   
\caption{Sample bird and drone sequences from the proposed dataset in upper and lower rows, respectively. Distant bird images are difficult to distinguish without using the sequence information. 
} 
\label{fig:motivation}
\end{figure}

\begin{figure*}[!ht]
\begin{center}
   \includegraphics[width=0.9\linewidth]{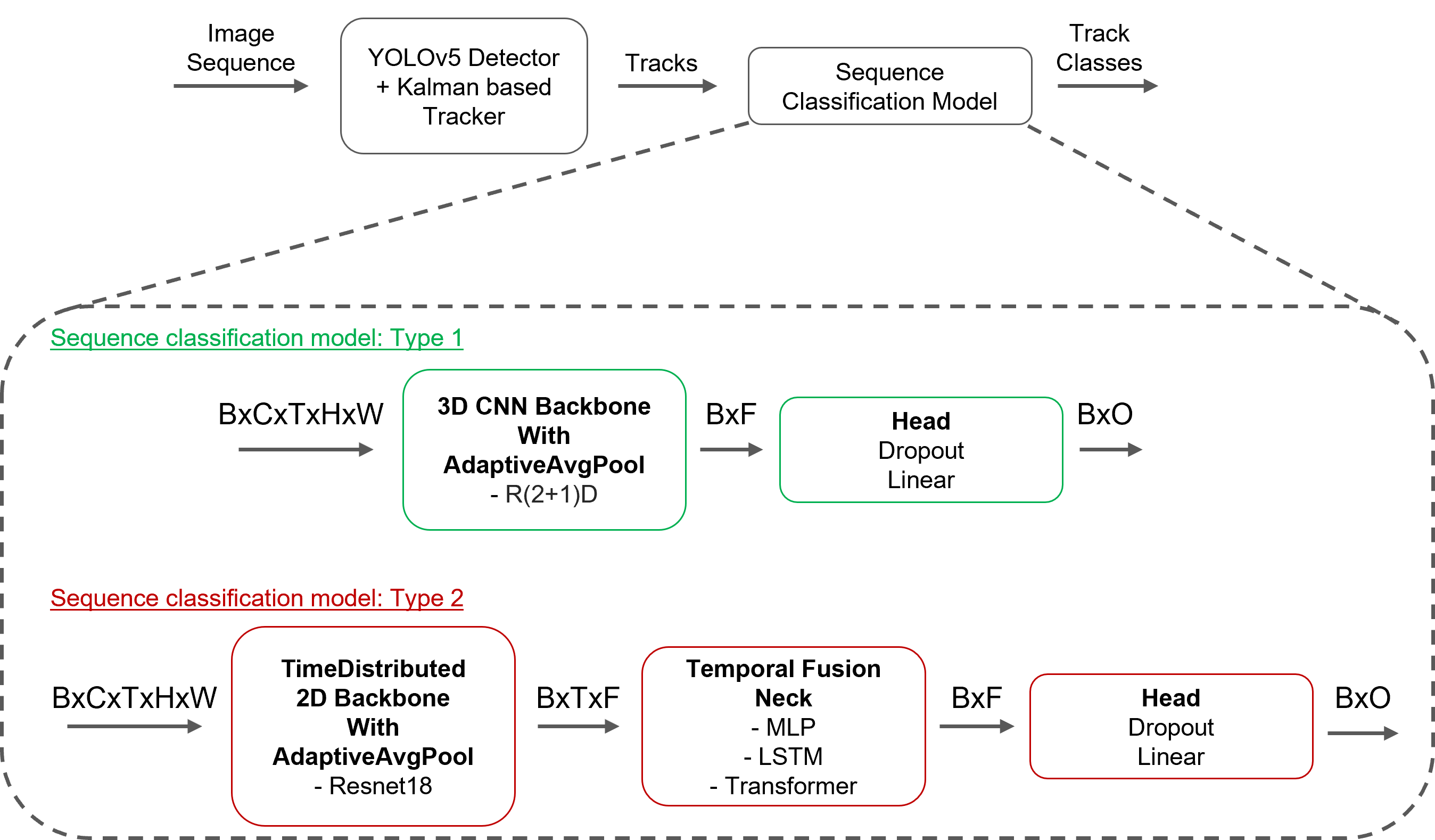}
\end{center}
   \caption{Overall diagram of the proposed work. B, C, T, H, W, F and O variables correspond to batch size, number of channels, number of timestamps, image height, image width, number of latent features, and number of output classes, respectively.}
\label{fig:overall}
\end{figure*}

Recently, an architecture based on an object detector and a Kalman-based tracker 
\cite{obss2021track} yielded the state-of-the-art results in WOSDETC2021 Drone vs Bird Challenge \cite{coluccia2021dvsb}. This method performs drone detection from a single frame without using the sequence information. An additional score boosting is applied to decrease the effect of false positives; however, it still fails to distinguish distant bird tracks from drone tracks in most scenarios. Because, as seen in Figure \ref{fig:motivation}, birds can easily be confused with drones at certain distances, and sequence information (i.e., wing movements) is crucial to distinguish distant bird images from drones. In order to overcome this issue, in this work, we extend the previous work of \cite{obss2021track}, with a sequence classification-based post-process stage, and aim at improving our false positive rates. To our knowledge, this work is the first to explore the addition of a video classification model to improve the drone vs. bird differentiation performance of an object detector.

We conduct experiments to explore 3D-CNN, LSTM and Transformer-based architectures for sequence-based classification. Moreover, we propose a new drone vs. bird video classification dataset to train and evaluate the proposed architectures.


\section{Related Work}


Video classification task has gained a significant success in the recent years. Specifically, the topic has gained more attention after the emergence of deep learning models as a successful tool for video classification. The reader may refer to \cite{rehman2021deep} for more information on the state-of-the-art on video classification literature.


Convolutional neural network based video classification methods can be analysed in 2 main categories: fully convolutional architectures and time-distributed 2D convolutional network based architectures.

As one of the fully convolutional video classifiers, R(2+1)D \cite{tran2018closer} proposes a ResNet \cite{he2016deep} inspired architecture with efficient alternative to 3D convolutions.
Similarly, SlowFast \cite{feichtenhofer2019slowfast} proposes a fully convolutional two stream model that extracts features from undersampled/upsampled versions of the input stream. Since these types of models are able to extract spatio-temporal features with a fully convolutional network, they are easy to implement and deploy. Thus, we have utilized R(2+1)D based architecture in our work.

Another alternative for video classification is to extract frame-based features from a CNN based backbone and feed to LSTM \cite{yue2015beyond}. Alternatively, LSTM can be replaced with Transformer encoder layer having multi-head self attention to extract temporal features \cite{abdali2021data}. These type of architectures can benefit from any pretrained image classification model. Thus, we explored the effect of these type of sequence models in our experiments.

\section{Proposed Work}
The proposed technique is based on our previous work in \cite{obss2021track} which utilizes a YOLOv5 \cite{yolov5} detector and a Kalman-based object tracker. Main focus of this work is to explore the effect of an additional video classification stage on top of tracker to improve the detection accuracy and false positives rates. Overall structure of the proposed technique can be seen in Figure \ref{fig:overall}.

\subsection{Video Classification Model}
As for the video classification stage, two types of architectures are utilized. First type includes the architectures consisting of fully convolutional backbones R(2+1)D \cite{tran2018closer} while the second type contains Resnet18  \cite{he2016deep} backbone combined with additional temporal feature extraction/fusion neck (LSTM/Transformer/MLP). Overall diagrams for Type 1 and Type 2 sequence classification models can be seen in Figure \ref{fig:overall}.

\subsection{Dataset}
To fine-tune and evaluate the proposed sequence classification modalities, a new drone vs bird video classification dataset (DvsB-Vid) has been formed from Drone vs Bird Detection Challenge \cite{coluccia2021dvsb} dataset which consists of a pool of 77 different video sequences containing drones, birds, planes on various backgrounds.

\begin{figure}[t]
\begin{center}
   \includegraphics[trim=210 50 150 20,clip=true,width=1\linewidth]{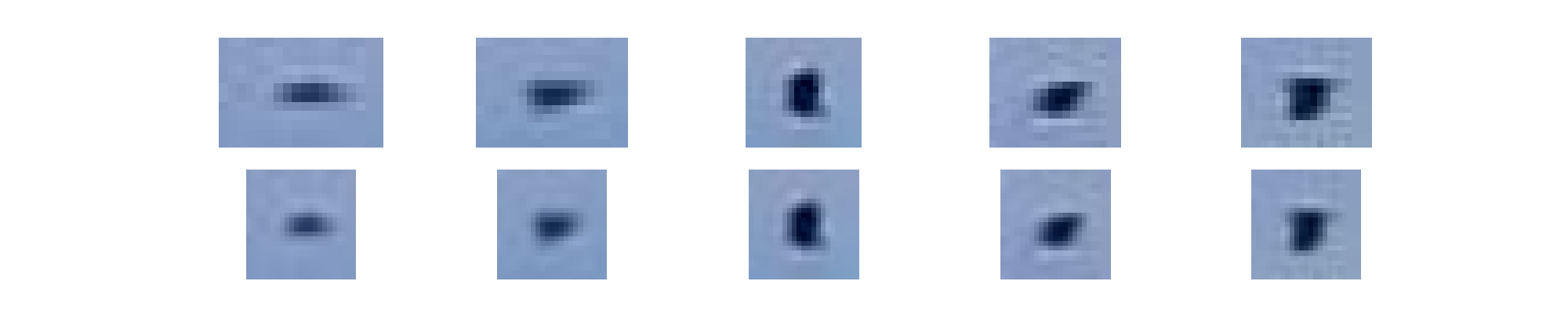}
\end{center}
   \caption{Sample bird sequence from created dataset. Upper row is the original track crops, below row is the resized version of it.}
\label{fig:bird}
\end{figure}

To create DvsB-Vid, we first applied the detection and tracking pipeline proposed in \cite{obss2021track} disregarding any track-based post-processing. Since the pipeline is trained without any negative labels, the resulting tracks contain many false positives containing bird, plane, and cloud objects. These track bounding boxes have been cropped and exported as video sequences and each sequence has been annotated as a drone or bird (disregarding other labels). While exporting a single track bounding boxes as frame sequences, after ten consecutive Kalman based tracker prediction that is not detected by the detector, a new frame sequence is started. Since all bounding boxes in a track sequence are not at a fixed size, all boxes in a track are resized to match the size of largest bounding box in the corresponding track. Dataset splits are the same as provided in \cite{isaac2021unmanned} for the original Drone vs. Bird Challenge dataset. There are 153 and 60 drone tracks/sequences are present in the training and validation split, respectively. Similarly, 92 and 20 bird tracks/sequences present in the training and validation split, respectively. In total, there are 48871 and 15526 frames in training and validation split, respectively.
Sample bird and drone sequence visuals are taken from the proposed dataset can be seen if Figures \ref{fig:bird} and \ref{fig:drone}, respectively.

\section{Results}

\begin{table*}[t]
    \centering
    \small
    \begin{tabular}{cccccccc}
    Architecture & Modality & Unfrozen & \# of Total & \# of Trainable & F1$_{drone}$ & F1$_{bird}$ & F1$_{macro}$ \\
    & & Backbone Block & Parameters & Parameters & & \\
    \hline
    ResNet18 & Single Image & 0 & \textbf{11.2M} & \textbf{1K} & \textbf{97.9} & 10.7 & 54.3\\

    R(2+1)D & Image Sequence & 0 & 31.3M & \textbf{1K} & 95.1 & \textbf{83.3} & \textbf{89.2}  \\
    ResNet18 + LSTM neck & Image Sequence & 0 & 11.4M & 181K & 93.2 & 79.0 & 86.1  \\
    ResNet18 + MLP neck & Image Sequence  & 0 & 11.4M & 262K & 90.2 & 51.9 & 71.0 \\
    ResNet18 + Transformer neck & Image Sequence  & 0 & 20.6M & 9.5M & 90.8 & 55.9 & 73.3 \\
    \hline
    ResNet18 & Single Image & 2 & \textbf{11.2M} & \textbf{10.5M} & \textbf{98.8} & 61.6 & 80.2 \\

    R(2+1)D & Image Sequence & 1 & 31.3M & 23.5M & \textbf{98.2} & \textbf{94.1} & \textbf{96.1}  \\
    ResNet18 + LSTM neck & Image Sequence & 2 & 11.4M & 10.7M & 96.1 & 87.4 & 91.8  \\
    ResNet18 + MLP neck & Image Sequence  & 2 & 11.4M & 10.8M & 90.3 & 55.7 & 73.0  \\
    ResNet18 + Transformer neck & Image Sequence  & 2 & 20.6M & 19.9M & 92.4 & 73.2 & 82.8 \\

    \hline
    
    \end{tabular}
    \caption{Category based and overall (macro) F1 scores for proposed modalities. Transfer learning and fine-tuning results are provided in upper and lower score result blocks, respectively.}
    \label{tab:results}
\end{table*}



\begin{figure}[t]
\begin{center}
   \includegraphics[trim=210 50 150 20,clip=true,width=1\linewidth]{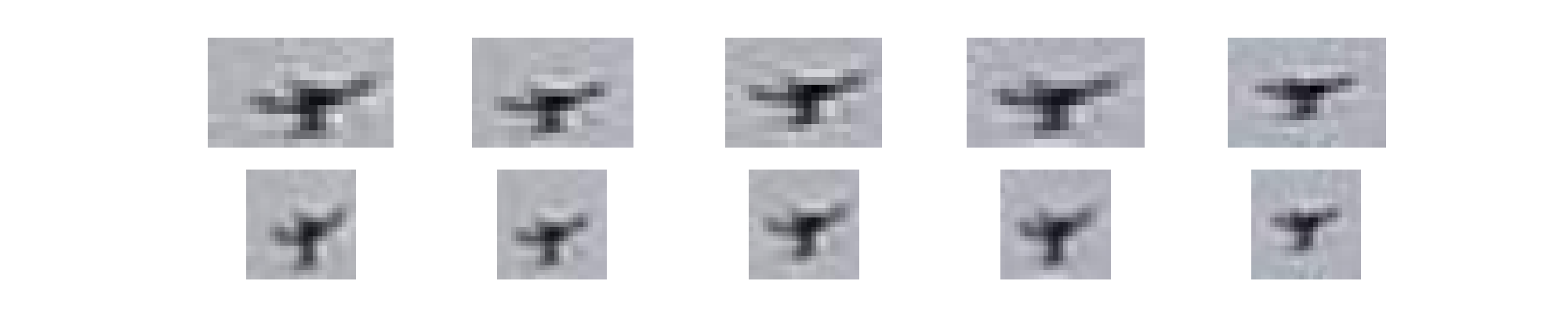}
\end{center}
   \caption{Sample drone sequence from created dataset. Upper row is the original track crops, below row is the resized version of it.}
\label{fig:drone}
\end{figure}

Comprehensive experiments have been performed to compare different architectures on the proposed DvsB-Vid dataset having drone and bird sequences. Sample bird and drone sequence visuals are taken from the proposed dataset can be seen in Figures \ref{fig:bird} and \ref{fig:drone}, respectively. In all experiments, frames are resized so that the short side is 224 pixels, and the same dataset is used in both video and image classification experiments (by only changing the data sampling collation pipelines).

In video classification experiments, R(2+1)D \cite{tran2018closer} backbone with Kinetics \cite{carreira2017quo} pretrained weights are used, 8 frames sequences are sampled randomly from 0.5 second clips, each batch is standardized with $mean=[0.45, 0.45, 0.45]$ and $std=[225, 0.225, 0.225]$, random crop, random short size resize between $[250,320]$ and random horizontal flip augmentations has been applied.

In image classification experiments, ResNet \cite{he2016deep} backbone with ImageNet \cite{russakovsky2015imagenet} pretrained weights are used, each batch is standardized with $mean=[0.485, 0.456, 0.406]$ and $std=[0.229, 0.224, 0.225]$, random crop and random horizontal flip augmentations has been applied.

For LSTM and MLP necks, hidden layer size and number of layers are selected as 64 and 2, respectively. For transformer neck, number of attention heads and number of layers are selected as 8 and 2, respectively.

For type 1 and type 2 sequence classification model experiments batch size is selected as 8 and 16, respectively. Moreover, batch size of 128 is used for image classification experiments.

After hyperparameter-tuning, optimizer is chosen as AdamW \cite{adamw} for image classification, LSTM and R(2+1)D based models and as SGD \cite{sgd} for MLP/Transformer based sequence classification models. In all experiments learning rate of 0.0001 and loss function of focal loss resulted in most stable results. Moreover, all experiments lasted 12 epochs with 1 full warm-up epoch and learning rate is decayed by 0.1 at 8th epoch.

All of the mentioned architectures are implemented using Pytorch framework \cite{pytorch}, pretrained weights are acquired from TorchVision framework \cite{torchvision}.

Two scenarios are considered in experiments for all proposed modalities: transfer learning and fine-tuning. In transfer learning scenario, all backbone layers are frozen and remaining parameters (neck and head) are learned from scratch. In fine-tuning scenario, most of the backbone parameters are unfrozen and updated during training. 

Category-based, overall (macro) F1 scores and the number of parameters for proposed modalities are given in Table \ref{tab:results}. Single image based classifier contains the least number of total parameters and yields the best drone classification score and the worst bird classification score for the transfer learning scenario. Sequence-based models yield better macro F1 in all cases except MLP neck based fine-tuning. LSTM based neck yields better performance compared to MLP which shows the importance of the sequential information extraction capability of RNN based layer. Transformer-based neck yields worse results than LSTM. This may be due to multi-head self-attention increasing the complexity significantly while training data is not that large. On a larger dataset, Transformer based head might yield better results. R(2+1)D based fully convolutional sequence model yields the best drone, bird and macro F1 scores in the fine-tuning scenario and the best bird and macro F1 scores in the transfer learning scenario. This is impressive considering in the transfer learning scenario, R(2+1)D based sequence model has only 1K trainable parameters. This can be explained with R(2+1)D having more pretrained parameters than other architectures.

\begin{figure}
     \centering
     \begin{subfigure}[b]{0.22\textwidth}
         \centering
         \includegraphics[width=\textwidth]{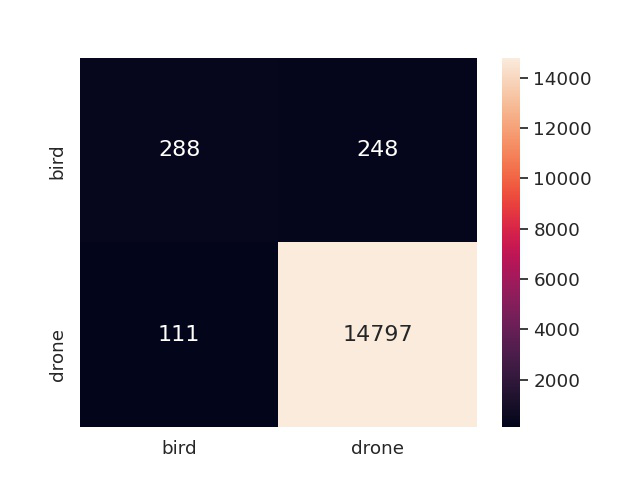}
         \caption{}
     \end{subfigure}
     \begin{subfigure}[b]{0.22\textwidth}
         \centering
         \includegraphics[width=\textwidth]{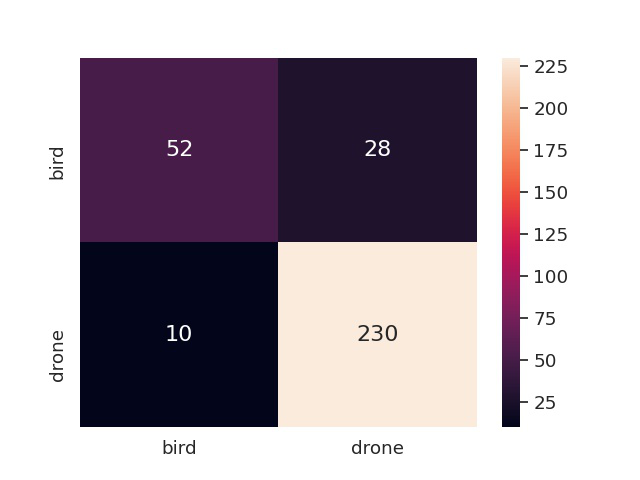}
         \caption{}
     \end{subfigure}
     \hfill
     \begin{subfigure}[b]{0.22\textwidth}
         \centering
         \includegraphics[width=\textwidth]{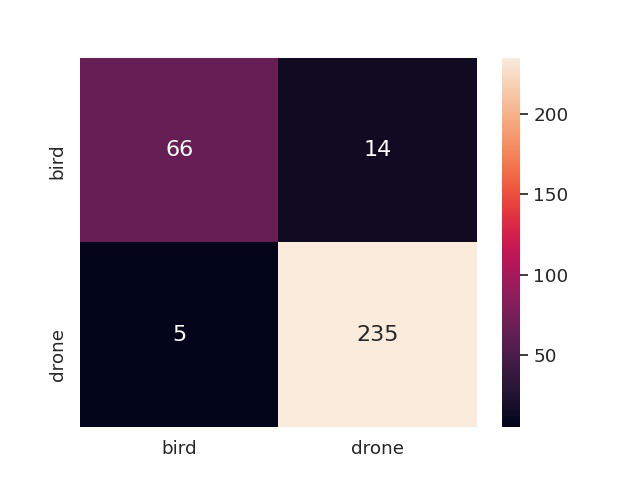}
         \caption{}
     \end{subfigure}
     \begin{subfigure}[b]{0.22\textwidth}
         \centering
         \includegraphics[width=\textwidth]{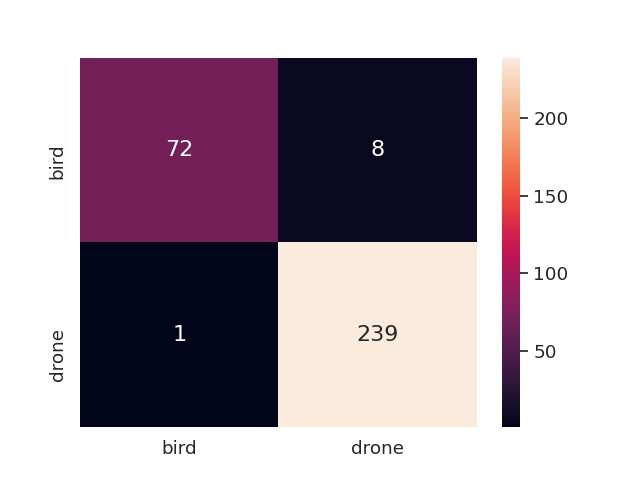}
         \caption{}
     \end{subfigure}
        \caption{Validation set confusion matrix plots of image classification, and Transformer/LSTM/R(2+1)D based sequence classification models in a, b, c and d respectively. Rows represent the true labels while columns represent the predicted labels.}
        \label{fig:conf}
\end{figure}

Validation set confusion matrices for some of the modalities in fine-tuning scenario can be seen in Figure \ref{fig:conf}. Poor bird classification result can be seen for image classification model while R(2+1)D based model provides very promising results.

\section{Conclusion}
In this work, the effect of a drone vs. bird classifier based post processing has been explored to decrease the number of false positives predicted by a drone detection and tracking architecture. LSTM/Transformer/MLP based temporal fusion necks are added on top of 2D and 3D image/video feature extractors to create sequence classification models. These sequence classifiers have been compared with single frame based image classifier on the newly proposed DvsB-Vid dataset. As experiments show, using sequence information, bird classification and overall scores can be increased by up to 73\% and 35\%, respectively. Among all sequence classification models, R(2+1)D based fully convolutional model yields the best transfer learning and fine-tuning results.

Recent image and video-based vision transformers can be utilized in future work to achieve better sequence classification results. Moreover, the sequence classification dataset can be enlarged by the addition of other noise labels which are not drone or bird.

\section*{Acknowledgement}
This research was done in relation to a project of OBSS Technology. We thank our colleagues from OBSS AI for their support, especially Kadir Sahin for assistance in annotating a portion of tracks, and Ogulcan Eryuksel for his valuable comments during development.

{\small
\bibliographystyle{ieee}
\bibliography{egbib}
}

\end{document}